\documentclass{article}
\usepackage{spconf,amsmath,graphicx}
\usepackage{mathtools, cuted}
\usepackage{float}
\floatstyle{plaintop}
\restylefloat{table}
\usepackage[tableposition=top]{caption}
\restylefloat{table}

\title{A Gaussian Scale Space Approach for Exudates Detection, Classification and Severity Prediction}
\name{Mrinal Haloi $^{1}$, Samarendra Dandapat $^{1}$, Rohit Sinha $^{1}$}
\address{$^{1}$Dept. Electronics and Communication Engineering, IIT Guwahati, (h.mrinal, samaren, rsinha)@iitg.ernet.in}

\begin{document}
%

\maketitle

\begin{abstract}
In the context of Computer Aided Diagnosis system for diabetic retinopathy, we present a novel method for detection of exudates and their classification for disease severity prediction. The method is based on Gaussian scale space based interest map and mathematical morphology. Iit makes use of support vector machine for classification and location information of the optic disc and the macula region for severity prediction. It can efficiently handle luminance variation and it is suitable for varied sized exudates. The method has been probed in publicly available DIARETDB1V2 and e-ophthaEX databases. For exudate detection the proposed method achieved a sensitivity of 96.54\% and prediction of 98.35\% in DIARETDB1V2 database.

\end{abstract}

\section{Introduction}
\label{sec:intro}
In recent days diabetic retinopathy (DR) is one of the severe eye diseases causing blindness. With early stage detection and treatment the patient can be saved from losing sight. Automatic computer aided diagnosis system will reduce burden on specialists. Also for monitoring and checking the progress of disease efficiently, automatic system will perform much better than human in terms of manual evaluation time. Since comparison and evaluation of images manually is a time consuming task and images are subject to various distortions. For accurate analysis of progress of diabetic retinopathy, detection of exudate is mandatory. Exudates are primary clinical symptoms of diabetic retinopathy. Two types of exudates namely soft exudate and hard exudate appear. Hard exudates are visible in non-proliferative diabetic retinopathy and soft exudates (cotton wool spots) in proliferative diabetic retinopathy. Hard exudates represent the accumulation of lipid in or under the retina secondary to vascular leakage and visible as discrete yellowish deposits in color fundus images. Cotton-wool spots are nerve fibre layer infarcts and they are visible as pale white rather than yellow. Also exudates are variable in sizes and shapes. A typical pathological retinal image is depicted in Fig. 1 to show features like the optic disc, the macula, the blood vessels and exudates.\\
\begin{figure}
  \centering
      \includegraphics[width=3.5in,height=2.6in]{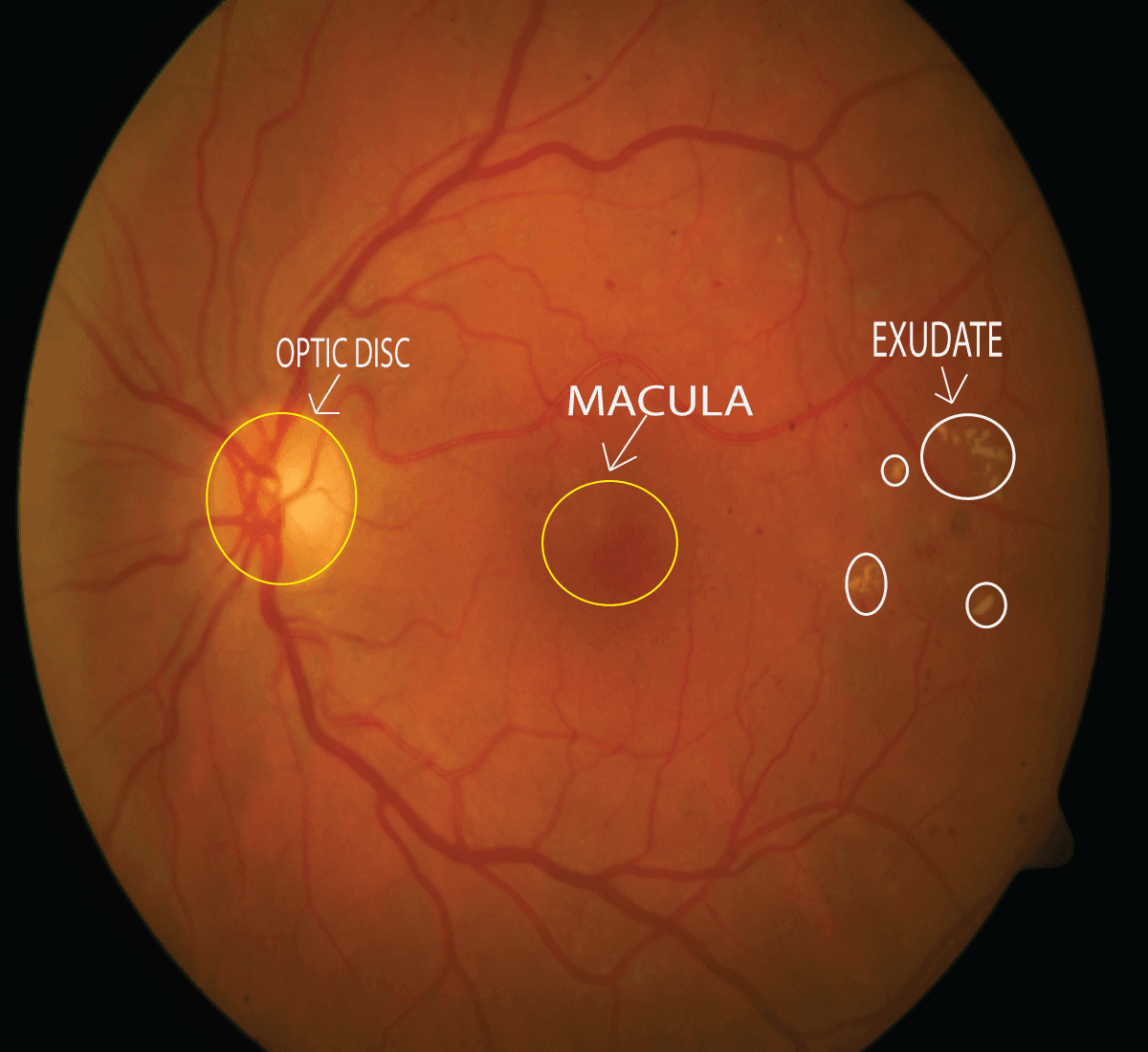}
\caption{Retinal Features}
\end{figure}
For identifying the stage of DR, classification of soft and hard exudates is foremost important to distinguish them for other retinal pathological features like drusen, heamorrhages, microaneurysms etc. Fundus images are prone to artifacts related to defocus, motion blur, fingerprints etc. This artifacts are prone to interpretation as pathological features because of their similarity with exudates and drusen. One important defect of fundus image is luminosity and contrast variation, which is improperly addressed in many exudate detection methods. 

Several methods have been presented for detection of exudates in colour fundus photograph using image processing and machine learning algorithms. Sanchez et al. \cite{c1} use a mixture model to separate exudate from background. An edge detection based method was used to remove other outliers. Giancardo et al. \cite{c2} use a Kirsch's edge method to assign score for exudate candidate to a pre-processed . They have used background estimation and image normalization for pre-processing. For classification and detection of drusen, exudates and cotton wool spots, a pixel wise classification algorithm is presented by Niemeijer et al. \cite{c3}. Zhang et al. \cite{c4} use mathematical morphological and contextual features for candidate extraction followed by random forest based classification method for exudate detection. Fuzzy c-means clustering based method was used for segmentation of features and neural network classifier for exudate detection by Osareh et al. \cite{c5}. Rocha et al. \cite{c15} have addressed the problem of detecting bright and red lession by using one novel algorithm, they made use of SURF features with machine learning method. But they failed to properly address the problems of luminance variation and artifacts.

In this work,we propose a novel exudate detection method using Gaussian scale space based interest map (GIMAP) and mathematical morphology. This approach is robust to artefacts and illuminance variation. Secondly a disease severity prediction method is developed by using information of exudate location with respect to the macula region and the optic disc. In addition to that we propose classification system using SVM for hard and soft exudate. Distinguishing exudate as hard and soft important for severity prediction also to identify type of diabetic retinopathy, whether it is non-proliferative or proliferative.
Section 2 presents our method. Experimental setup and the results are discussed in section 3 and finally conclusions are are drawn in section 4.

\section{Method}
\subsection{Exudate Detection}
\noindent
Exudate are the bright lessons found in retinal image, caused due to diabetic retinopathy, a most common disorder of eye with patient having diabetes. It is also a main reason of blindness. For detection of the optic disc and fovea, the method described by Niemeijer et al. \cite{c6} was used. Steps involve in this method is shown in Fig. 2. 
\begin{figure}
  \centering
      \includegraphics[width=3.2in,height=4.0in]{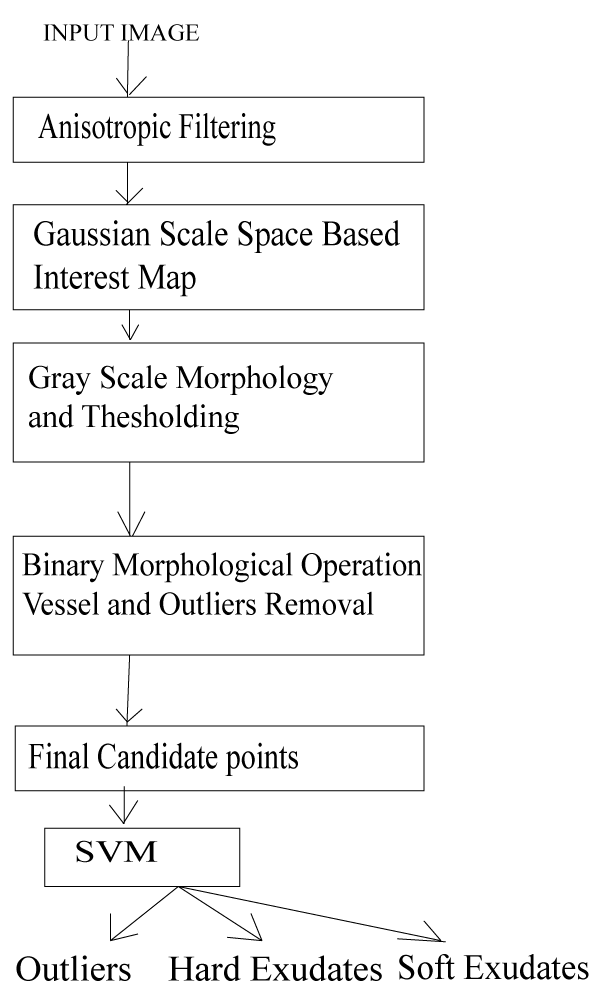}
\caption{Method Overview}
\end{figure}
\subsubsection{Preprocessing}
For correct detection and classification of hard and soft exudates it is important to reduce noise while preserving edges. In this method anisotropic diffusion filtering \cite{c7} was used for reducing noise in the images and preserving edges. Anisotropic diffusion filters have been proved to be succesful in edge preserving smooting and denoising of medical images. Hard and soft exudates possesses discriminative edge structures, preserving this edge while reducing noise is also one of important step towards classification. Step involves in filtering is as follows \cite{c7}.
 
\begin{equation}
\frac{\partial}{\partial \sigma}I(r,g,b,\sigma) = \nabla \bullet (c(r,g,b,\sigma)\nabla I(r,g,b,\sigma))
\end{equation}
\begin{equation}
c(r,g,b,\sigma) = f(|\nabla I(r,g,b,\sigma)|)
\end{equation}
\begin{equation}
c(r,g,b,\sigma) = \frac{1}{1 + (\frac{|\nabla I(r,g,b,\sigma)|}{K})^{1 + \alpha}} , \alpha > 0
\end{equation}
where I is the input image, (r,g,b) is color channel of it, K is diffusion constant and $\sigma$ is noise standard deviation over which algorithm will be iterated to find solution. The diffusion function $c(r,g,b,\sigma)$ is a monotonically decreasing function of the image gradient magnitude.

\subsubsection{Gaussian Scale Space Construction}
Gaussian Scale space based interest map can capture local image structure and scale invariant image features. Since exudate varies on the basis of size and hence different exudate will response to different scales of Gaussian. For identifying all exudate interest map using Gaussian scale space was constructed \cite{c8}. First step in the construction of GIMAP is the computation of 1st derivative of Gaussian as filters at several scales and smoothed the derivative using Gaussian filter and take absolute values of derivatives. In addition to that Laplacian of Gaussian for each scale was computed. Because of colour difference between hard and soft exudates, they respond variably to colour channels. Now taking the maximum response of all the colour channels final interest map for the scale is constructed. This process is repeated over different scales. For each scale we have two filter 1st derivative Gaussian and Laplacian of Gaussian, take maximum of absolute values from both filter ouputs in the interest map. Since our concerned features may appear in variable scales so interest map of each individual map is combined using maximum operation to form a decision making interest map. Scale used in these works are $ (\sqrt[]{2},k\sqrt[]{2} ... k^{n}\sqrt[]{2})$. The decision making interest map also contain some outliers such blood vessels, vein, haemorrhage etc. Filtering process is described by following equations \cite{c9}.

\begin{equation}
G_{\sigma}(x,y) = \frac{1}{2\pi \sigma^{2}} e^{ - \frac{x^2 + y^2}{2\sigma^2}}
\end{equation}
\begin{equation}
\begin{aligned}
\frac{\partial G_{\sigma}(x,y)}{\partial x} \propto xe^{ - \frac{x^2 + y^2}{2\sigma^2}}\\
 \frac{\partial G_{\sigma}(x,y)}{\partial y} \propto ye^{ - \frac{x^2 + y^2}{2\sigma^2}}
\end{aligned}
\end{equation}
\begin{equation}
\begin{aligned}
\nabla^2 G_{\sigma}(x,y) = \frac{\partial^2 G_{\sigma}(x,y)}{\partial^2 x} + \frac{\partial^2 G_{\sigma}(x,y)}{\partial^2 y} \\
\nabla^2 G_{\sigma}(x,y) \propto G(x,y,k^2\sigma) - G
(x,y,\sigma)  
\end{aligned}
\end{equation}
 For computation of Laplacian of gaussian we have used difference of Gaussian approximation.
If $I(v,k,f)$ be a filtered image, $v[r,g,b]$ is its three color channel at a scale $k$ using filter $f$, then the interest map for this scale is obtained by selecting maximum response over color channel as follows.
\begin{equation}
I(k,f) = max_{v}I(v,k,f) 
\end{equation}
And the decision making interest map is obtained taking maximum over filter $f$ and then scale $k$ as follows.
\begin{equation}
dMap = max_{k} max_{f}I(k,f)
\end{equation}

\subsubsection{Binarization and Vessel Removal}
The decision map obtained from above procedure need to be improved by reducing outliers present. First step is to enhance the features by using grayscale morphological operations. Here closing operation with a 'disk' structuring element of size 2 and 3 was used. 
\begin{equation}
\begin{aligned}
Dilation: (f \oplus B)(x,y) = max{f(x-s, y-t)|(s,t) \in B}\\
Erosion: (f \ominus B)(x,y) = max{f(x+s, y+t)|(s,t) \in B}
\end{aligned}
\end{equation}
\begin{equation}
\begin{aligned}
Opening:  (f \circ B)  = (f \ominus B) \oplus B  \\
Closing: (f \bullet B)  = (f \oplus B) \ominus B
\end{aligned}
\end{equation}

If $iMap1$ and $iMap2$ are obtained after closing operation, then resulting enhanced image is obatined by following operation.
\begin{equation}
iMap = max_{i,j}(iMap1, iMap2)
\end{equation}
where $(i, j)$ are pixels position in both images.

In the second step convert the interest map to binary map. For binarization of the interest map Sauvola's \cite{c10} local adaptive thresholding technique is used. Local threshholding is efficient in this particular situation beacuse pixels values of hard and soft exudates vary significantly and to get both of those in final map we need to binarize using local windows. Let $I(x,y)$ is interest map image, take a window size of $9\times9$, $m_{(x,y)}$ and $\sigma_{(x,y)}$ be the mean and standard deviation of window cantered at $(x,y)$, then threshold $th(x,y)$ will be defined as follows.

\begin{equation}
th(x,y) = m_{(x,y)}[1 + c(\frac{\sigma_{(x,y)}}{\Sigma} - 1)]
\end{equation}

 where $\Sigma$ is maximum of standard deviation of all windows and $c \in [0.2 0.5]$ is a parameter.

Along with exudate other features such as blood vessels and veins is also detected due to Gaussian scale space. Morphological opening operation and connected component analysis was used for removing those unwanted features. Blood vessels and veins are characterized by thin long geometrical structures. A rectangle structural element having width and length larger than the thickness of those blood vessels is used for removing those structures. Remaining vessels and outliers left after this operation will be removed by using concept of convexity and connected component analysis. For each connected region we will compute convex hull of the region and define the following $probReg$ eq. (13) term.
\begin{equation}
probReg = \frac{\Sigma_{i,j}CH}{\Sigma_{i,j}R}
\end{equation}
 The $probReg$ term computes probability of a connected region being exudate or not. It is the ratio of the area of the region convex hull and its area. This term will give approximate idea of the connected region. If the value of $probReg > 0.8$ then the region will be discarded, the connected region will be considered as vessel structures or flash artifacts. 

where $CH$ is convex hull of the region $R$ and $(i,j)$ denotes the pixel positions. 
If convexity of a connected region is less than 0.2 then the area will be discarded. 
Also we want to ensure that artifacts like flares gets removed from GIMAP. From observation it has been noticed that flares are circular shape object. None of the features like hard and soft exudate have shapes like that of flares, approximate circular. Compactness ($C$) is defined below eq. (14) of  every connected region of the GIMAP should be less than $\frac{\pi}{5}$. 
\begin{equation}
C = \frac{A_{R}}{P^{2}_{R}}
\end{equation} 
where $A_{R}$ and $P^{2}_{R}$ are area and perimeter of the region $R$ respectively.
In the final step we will use colour and luminance features to select exudates and remove outliers, in general we didn't observe outliers like flares, flash etc after adapting previos mentioned steps. But still there are possibility fo including drusens. Hard and soft exudates are characterized by its yellowish and whitish color significantly different from that of haemorhages, microanerysms and other related features. At the end, a support vector machine classifier is used to get final labels of hard and soft exudates.

\subsection{Classification of Exudates using SVM}
Two types of exudates namely hard and soft are associated with different diabetic retinopathy symptoms. We will use a Support Vector Machine classifier to differentiate between soft and hard exudates and outliers. SVM is one of most effective classifier, with its good generalization ability. Selection and extraction of feature vector is one of the challenging task for efficient classification. Since hard and soft exudates are varied in color magnitude we will use mean R, G, B values as feature for each detected regions. In addition, to ensure luminance invariance in classification process, images will be converted to Lab color space and values of a, b channels will be used as features. Both hard and soft exudates are varied in terms of shape and edge structures. Hard exudates are chracterizes by uneven edge and soft exudates are by smooth edge and circular structures. Firstly we will design features vector for classification by using following data. Tabel 1 shows the features used in this method for SVM.

\begin{table}[H]
\caption{Features for SVM}
  \begin{tabular}{*{20}{c}}
\hline
Feature & Description\\
\hline
1 & Area of Connected Region \\
\hline
2-4 & Mean R, G, B Intensity Values \\
\hline
5-7 & Standard deviation of R, G, B intensity values \\
\hline
8-9 & Mean a, b intensty values in Lab color space \\
\hline
10-11 & Standard deviation of a,b intensity values \\
\hline
12 & Ecentricity of the region \\
\hline
13 & Extent of the region \\
\hline
14-15 & Major and Minor axis length\\
\hline
16 & Convexity \\
\hline
17 & Gradient values of edge pixels of the region \\
\hline
18 & Compactness of the Region \\
\hline
19 & Energy of the Region \\
\hline
20-22 & Color contrast with neighbour regions \\
\hline
\end{tabular} 
  
\end{table}

Features vector constructed from above ideas is 22 dimensional. For SVM classification, we have three class specifically hard exudate, soft exudate and outliers. System is trained with hard, soft exudates and non exudates pixels data including artifacts from expert labeled images.


\begin{figure}
  \centering
      \includegraphics[width=3.5in,height=2.2in]{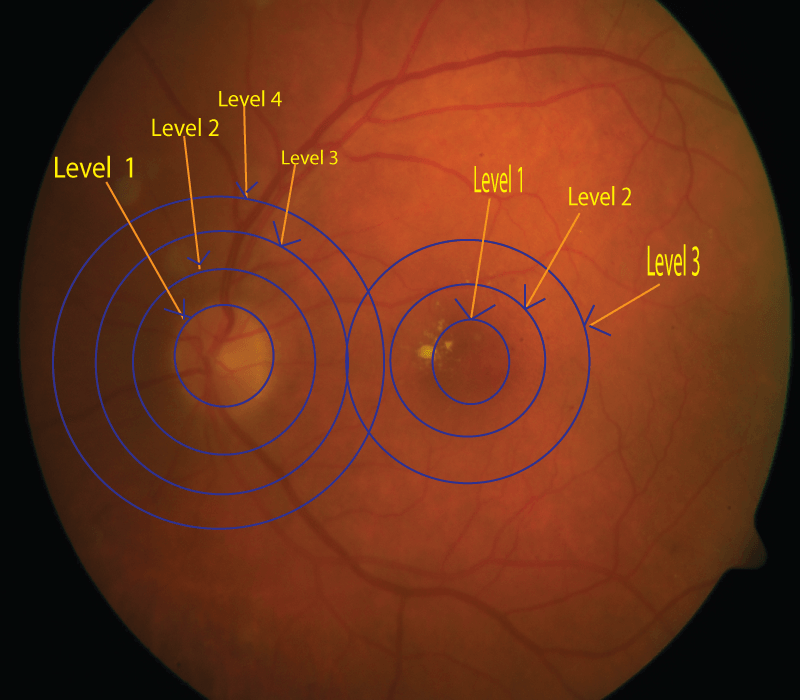}
\caption{Severity level Groups}
\end{figure}

\subsection{Severity Detection}
As per ophthalmologists, exudate (or other retinal pathology such as microaneurysms, heamorrhages, drusen etc.) location with respect to the fovea and the optic disc determine severity of the diseases. If exudates location are close to those features and causes defect to those features then patients may lose vision. As per the norm of the International Council of Ophthalmology Diabetic Retinopathy \cite{c14} divided in five groups specifically: none, mild, moderate, severe, and proliferate. We will define some rule to detect severity by using the fovea and the optic disc location as centre of two set of concentric circles. Radius of each concentric circles with the fovea location as centre is multiple of 80 pixels. The radius selection for each circle based on the average value of the optic disc width and height. Also for each concentric circles with the optic disc centre location as circle centre is multiple of 55 pixels. This setting valid for images with width 1500 pixels and height 1152 pixels. If exudates reside in vey inner circle then we label it as $Level_{1}$, other consecutive circle will be known as $Level_{ N}$, where N is the index number of the circle, bigger N means circle is of bigger radius. For each circle area of exudate(No of pixels) will be calculated. Finally by using area and location information diseases severity will be predicted. If location of exudate in $Level_{1}$ circle then on the basis of its areas it will included in proliferate or severe groups. Here area of $Level_{N}$ circle means area of $(Level_{N} - Level_{N-1})$ circle. Fig. 3 depicts the scenarios described above. 

\begin{equation}
\begin{aligned}
Severity \propto (c_{1}*A + c_{2} * \frac{1}{D}) \\
c_{1} = f(D) \\
c_{2} = f(D)
\end{aligned}
\end{equation}
where $A$ denotes area of exudate region, $D$ its distance from either the optic disc center or the fovea and $c_{1}, c_{2} \in [0, 1]$ are two constant function of distance.
\begin{equation}
Proliferate: x_{i} \in C_{1} \cap \Sigma_{i}x_{i} > \frac{1}{16}A_{C_{1}}
\end{equation}
\begin{equation}
\begin{split}
Severe: (x_{i} \in C_{1} \cap \Sigma_{i}x_{i} \leq \frac{1}{16}A_{C_{1}}) \cup\\
 (x_{i} \in C_{2} \cap \Sigma_{i}x_{i} > \frac{1}{16}A_{C_{2}})
\end{split}
\end{equation}
\begin{equation}
\begin{split}
Moderate: (x_{i} \in C_{2} \cap \Sigma_{i}x_{i} \leq \frac{1}{16}A_{C_{2}}) \cup \\
 (x_{i} \in C_{2} \cap \Sigma_{i}x_{i} > \frac{1}{16}A_{C_{3}})
\end{split}
\end{equation}
\begin{equation}
\begin{split}
 Mild: (x_{i} \in C_{3} \cap \Sigma_{i}x_{i} \leq \frac{1}{16}A_{C_{3}}) \cup \\
(x_{i} \in C_{2} \cap \Sigma_{i}x_{i} > \frac{1}{16}A_{C_{4}})
\end{split}
\end{equation}

where $x_{i}$ is a exudate pixel, $C_{i}$ denotes $Level_{i}$ circle and $A_{C_{i}}$ its corresponding area.

\begin{figure}
  \centering
      \includegraphics[width=3.4in,height=4in]{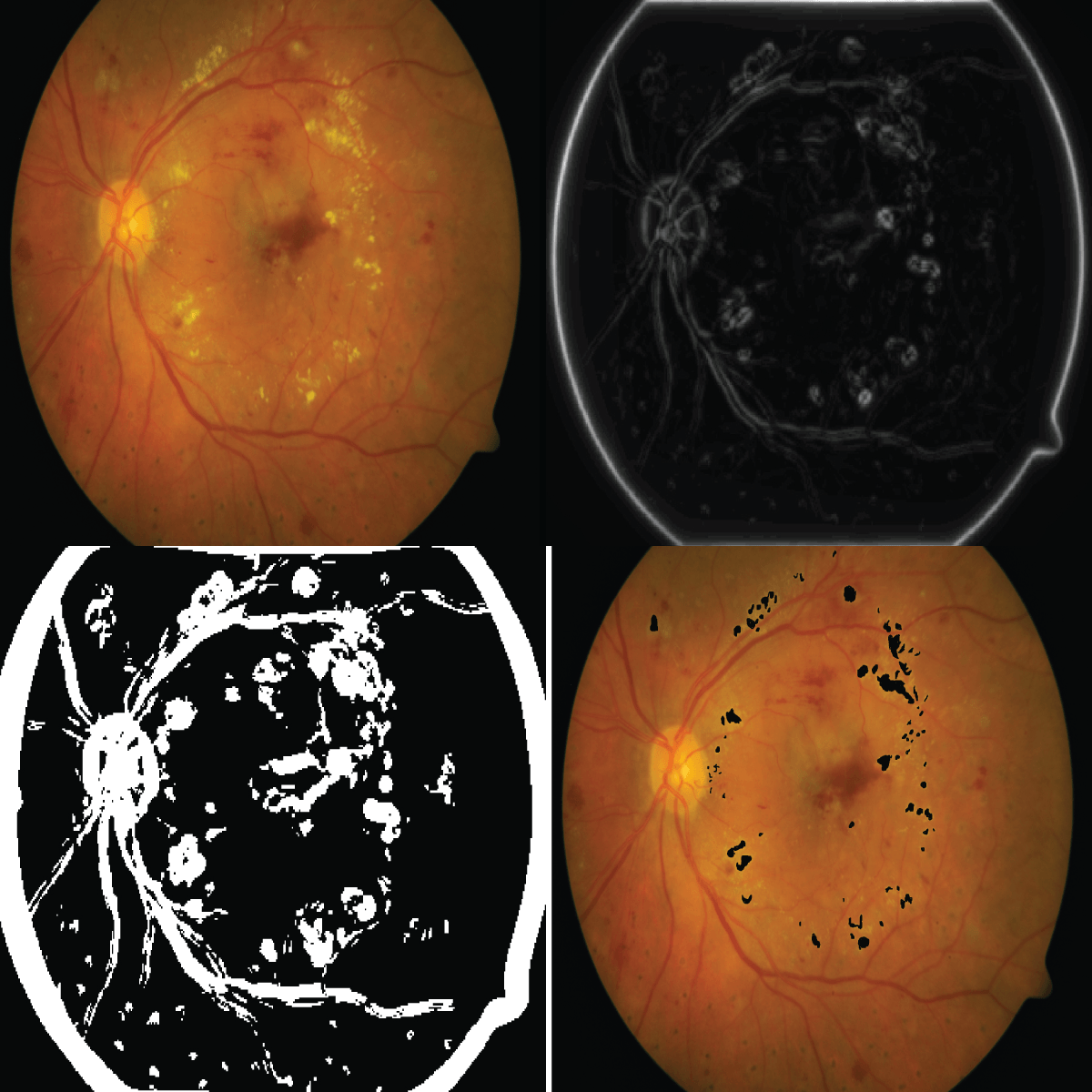}
\caption{Left: Original Image, Right: GIMAP, Down - Left: After Binarization, Down -  Right: Final detected Exudate}
\end{figure}

\begin{figure}
  \centering
      \includegraphics[width=3.3in,height=2.0in]{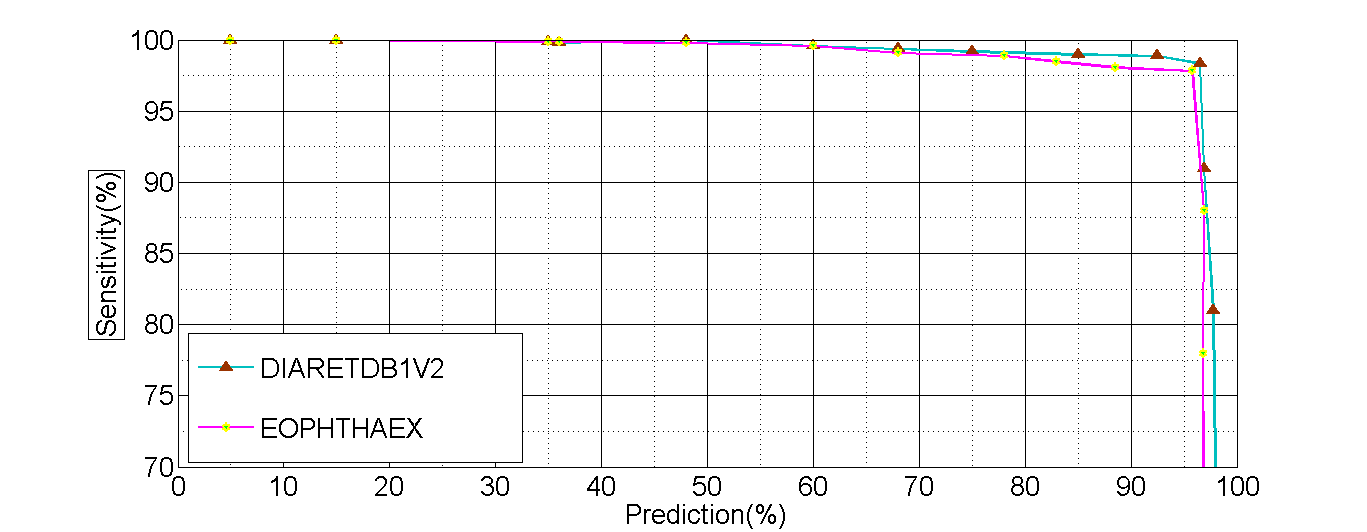}
\caption{Sensitivity and prediction variation }
\end{figure}

\section{RESULT and DISCUSSIONS}
\noindent
For analysing the accuracy of our method on different set of images, we have used publicly available DIARETDB1 \cite{c11} and e-ophthaEX \cite{c12} dataset. Images with various difficulty from these databases have been chosen for testing our algorithm accuracy, this dataset include image with pathological features such as haemorrhages, exudates and microaneurysms. This dataset provided by experts in Ophthalmology with proper pixel wise annotation of features location in images. This three dataset includes 400 retinal images with variety of pathological symptoms. For implemetation MATLAB platform in a windows 8.1 machine with Intel i7 processor was used.

For detection of exudate Gaussian scale space was constructed using 10 different scale. At first each images is resized by a factor of $ \frac{400}{max(rownum, colnum)}$ using cubic interpolation method for reducing computational time. For each scale first derivative of Gaussian and Laplacian of Gaussian is computed for analysing structure of exudates present in retinal images. By using those values a decision making interest map is formed. Different types of outliers were removed using morphological connected component analysis and opening operation.

For accuracy analysis of exudates detection we will compute true positive (TP) a number of exudates pixels correctly detected, false positive (FP) a number of non-exudate pixels which are detected wrongly as exudate pixels, false negative (FN) number of exudate pixels that were not detected and true negative (TN) a number of no exudates pixels which were correctly identified as non-exudate pixels. Also sensitivity and specificity at pixel level is computed. Thus the global sensivity SE and the global specificty SP and accuracy AC for each image is defined as follows.

\begin{equation}
\begin{aligned}
SE = \frac{TP}{TP + FN}\\
PRED = \frac{TP}{TP + FP}\\
SP = \frac{TN}{TN + FP} \\
AC = \frac{TP + TN}{TP + TN+FP + FN}
\end{aligned}
\end{equation}

A detailed result of accuracy obtained is illustrated in Table 1.
\begin{table}[H]
  \begin{tabular}{p{0.22\linewidth}p{0.21\linewidth}p{0.13\linewidth}p{0.13\linewidth}p{0.10\linewidth}}
\hline
Database & Resolution & Sensitivity & Prediction & AUC\\
\hline
eophthaEX & $ 960 \times 1440 $ & 95.82 & 97.85 & 0.962\\  
\hline
DIARETDB1v2 & $ 1152 \times 1500 $ & 96.54 & 98.35 & 0.968\\  
\hline
\end{tabular} 
  \caption{Result of  Exudate detection}
\end{table}

\begin{table*}[t]
  \centering
  \begin{tabular}{*{20}{c}}
\hline
Database & Resolution & \# images & Type & True Positive\/Per image & False Positive\/Per image\\
\hline
eophthaEX & $ 960 \times 1440 $ & 80 & Hard Exudate & 98.73 \% & 3.05 \% \\  
\hline
DIARETDB1v2 & $ 1152 \times 1500$ & 80 & Hard Exudate &  98.51 \% & 2.16 \% \\  
\hline
DIARETDB1v2 & $ 1152 \times 1500 $  & 80 & Soft Exudate & 98.23 \% & 2.35 \% \\  
\hline
\end{tabular} 
  \caption{Result of  Exudate Classification}
\end{table*}
Fig. 4 shows exudate detection intermediate and final step results. First row shows the original image and its corresponding GIMAP, where as in second row image obtained after binarization and final detection result is depicted. From this Figure it can be noticed that GIMAP respones to retinal features such as the optic disc, the blood vessels etc and performance of postprocessing steps are well observed from final detection result.  

Tabel 2 depicts the result obtained on 89 image each of DiARETDB1V2 nad EOPHTHAAEX databses. Computation of sensitivity and prediction are pixel based. Value of TP, FP, TN, FN are measured as number of pixels and eq.(19) is used for sensitivity and prediction calculation. From the Table 2 it can be clearly observed that the presented method is robust to other databases also. 
Fig. 5 shows the variation of sensitivity vs prediction on the DIARETDB1V2 and EOPhTHAEX database. Performance on DIARETDB1V2  with sensitivity of 96.54\% and prediction of 98.35\% is slightly better than on EOPHTHAEX with sensitivity of 95.82\% and prediction of 97.85\%. From this it can be infered that this method is almost dataset independent.

According to Fig. 6, on the DIARETDB1V2 datbase AUC value of this method is 0.968 with sensitivty of 96.54\%. This result is satisfactory for practical purpose of mass screening. A comparisons of sensitivity vs 1-specificity with recent state-of-the-art  method is explored in Fig. 6. The method proposed by Zhang et al.[4], achieved AUC value of 0.95 in the same database. Also the proposed by Giancardo et al.[2], achieved AUC value of 0.87. 

Tabel 4 depicts sensitivity, specificity and AUC comparisons with other methods. Eventhough not all these methods have used a common dataset, but this comparions is to show the advantages of this method in terms of those measurements. 
 A comparison of AUC values with latest state-of-the-art methods is shown in Table 5. Our method perform very well in comparison to existing methods.

\begin{figure}
  \centering
      \includegraphics[width=3.3in,height=2.0in]{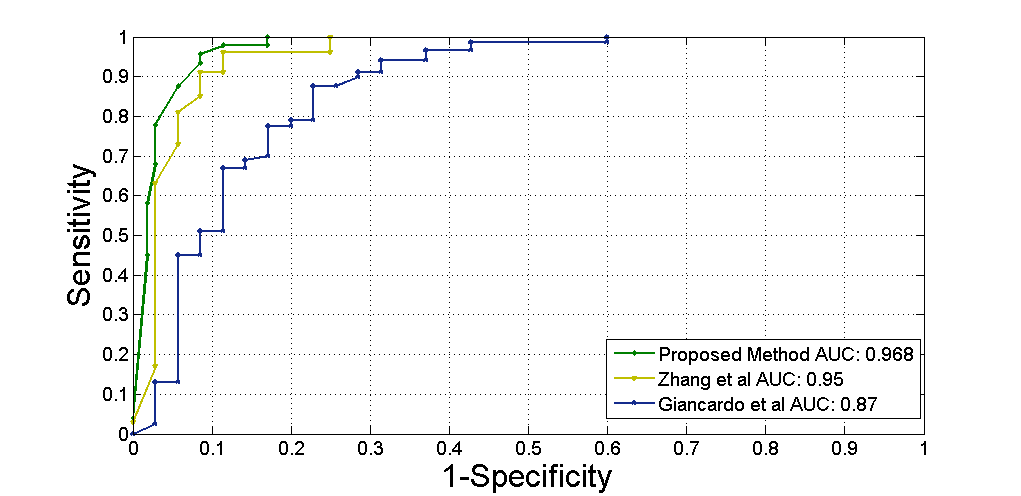}
\caption{Sensitivity and 1-specificity variation comparions with existing methods on eophthaEX}
\end{figure}

For classification of soft and hard exudates we have used LibSVM \cite{c13} package. Each image also converted to Lab colour space for getting exudate luminance and colour features separately. Training data is a collection of total 5000 data points having 14 dimension. Training data used both DIARETDBV2 and EOPHTHAEX dataset. For estimating generalisation error a cross validation (10-fold) method was used, where training set was divided into 1:9 parts. For each time 9 parts will be used for training and remaining one part for testing. Separate evaluation was done on both dataset for soft and hard exudate classification. In Table 3 detailed result of classification in pixel level is shown. This final result is very satisfactory for practical use of  mass screening purpose. \\
One of the foremost advantages of our method is that we don't need to detect the optic disc or the vessels separately to segment out exudates, as this method automatically remove those features in detection process. Most of the already exised method \cite{c4} need separate detection for the optic disc and the blood vessels in the exudate detection phase. 

\begin{table}[H]
  \centering
  \begin{tabular}{*{20}{c}}
\hline
Methods & Sen (\%) & Spec (\%) & AUC \\
\hline
Proposed Method & 96.54 & 93.15 & 0.968 \\
\hline
Zhang et al. \cite{c4} & 96.0 & 89.0 & 0.95 \\ 
\hline
Niemeijer et al. \cite{c3} & 95.0 & 86.0 & 0.95 \\
\hline
Sinthanayothin et al. \cite{c16} & 88.5 & 99.6 & N\/A \\
\hline
Osareh et al. \cite{c5} & 93 & 94.1 & N\/A \\
\hline
Walter et al. \cite{c17} & 92.74 & 100 & N\/A \\
\hline
\end{tabular} 
  \caption{Comparison results for Exudate detection, Dataset are not same}
\end{table}

\begin{table}[H]
  \centering
  \begin{tabular}{p{0.24\linewidth}p{0.13\linewidth}p{0.15\linewidth}p{0.22\linewidth}}
\hline
Database &Proposed Method   & Zhang et al.(2014) & Giancardo et al(2012)\\
\hline
eophthaEX & 0.952  & 0.930  & 0.900   \\  
\hline
DIARETDB1v2 & 0.968 & 0.950 &0. 930  \\  
\hline
\end{tabular} 
  \caption{Comparison of AUC on two standard database}
\end{table}

\section{Conclusion}
A novel method for computer aided diagnosis of retinal image for exudate detection and analysis is proposed. We have got considerable accuracy over two different dataset comprised of several varieties of images specifically illuminance changes, with other pathological features etc. Also our machine learning based classifier SVM works very well for exudate classification with designed features. This system can be used for automated processing of pathological images related to diabetic retinopathy, also will be very effective for mass screening. In near future, we will incorporate microaneurysms and haemorrhages detection to the system to enhance its credibility to evaluate the degree of diabetic retinopathy

\end{document}